\def\year{2021}\relax
\documentclass[letterpaper]{article} % DO NOT CHANGE THIS
\usepackage{aaai21}  % DO NOT CHANGE THIS
\usepackage{times}  % DO NOT CHANGE THIS
\usepackage{helvet} % DO NOT CHANGE THIS
\usepackage{courier}  % DO NOT CHANGE THIS
\usepackage[hyphens]{url}  % DO NOT CHANGE THIS
\usepackage{graphicx} % DO NOT CHANGE THIS
\usepackage{natbib}  % DO NOT CHANGE THIS AND DO NOT ADD ANY OPTIONS TO IT
\usepackage{caption} % DO NOT CHANGE THIS AND DO NOT ADD ANY OPTIONS TO IT
\usepackage[utf8]{inputenc} % allow utf-8 input
\usepackage[T1]{fontenc}    % use 8-bit T1 fonts
\usepackage{microtype}      % microtypography

\usepackage{subcaption}
\usepackage{wrapfig}
\usepackage{float}
\usepackage{chngpage}
\usepackage{enumitem}
\usepackage[dvipsnames]{xcolor}

\usepackage{tikz}
\tikzset{every picture/.style={line width=1pt}} %
\usepackage{graphicx}
\usepackage{amsmath}
\usepackage[
  pagebackref,
  breaklinks=true,
  pageanchor=true,
  plainpages=false,
  pdfpagelabels,
  bookmarks=false,
  bookmarksnumbered,
  colorlinks=true,
  linkcolor=ForestGreen,
  citecolor=MidnightBlue,
  urlcolor=magenta,
]{hyperref}
\usepackage{mathtools}
\usepackage{microtype}
\usepackage{url}
\usepackage{amsfonts}
\usepackage{amssymb}
\usepackage{amsthm}
\usepackage{mathrsfs}
\usepackage{verbatim}
\usepackage{booktabs}
\usepackage{siunitx}
\usepackage{multirow}
\usepackage{footnote}
\usepackage{blindtext}
\usepackage{mwe}
\usepackage{adjustbox}
\usepackage{algorithm}
\usepackage[linesnumbered,ruled,algo2e]{algorithm2e}
\usepackage[font=small,labelfont=bf]{caption}

\usepackage[toc,page]{appendix}
\usepackage{comment}

\usepackage{amsmath,amsfonts,bm}

\def\eqref#1{equation~\ref{#1}}
\def\1{\bm{1}}

\def\rvr{{\mathbf{r}}}

\DeclareMathAlphabet{\mathsfit}{\encodingdefault}{\sfdefault}{m}{sl}
\SetMathAlphabet{\mathsfit}{bold}{\encodingdefault}{\sfdefault}{bx}{n}

\newcommand{\E}{\mathbb{E}}

\newcommand{\abs}[1]{\mathopen{}\left| #1 \mathopen{}\right|}
\newcommand{\defeq}{\mathrel{\stackrel{\makebox[0pt]{\mbox{\normalfont\scalebox{.5}{$\triangle$}}}}{=}}}

\newcommand{\methodname}{\textit{AutoDropout}}
\title{\methodname: Learning Dropout Patterns to Regularize Deep Networks}
\author {
  Hieu Pham\textsuperscript{\rm 1,2} \textnormal{and} 
  Quoc V. Le\textsuperscript{\rm 1} \\
}
\affiliations {
  \textsuperscript{\rm 1} Google Research, Brain Team, Mountain View, CA 94043 \\
  \textsuperscript{\rm 2} Carnegie Mellon University, Pittsburgh, PA 15213 \\
  \texttt{\{hyhieu,qvl\}}@google.com
}

\begin{document}

\maketitle

\begin{abstract}
Neural networks are often over-parameterized and hence benefit from aggressive regularization. Conventional regularization methods, such as Dropout~\citep{dropout} or weight decay, do not leverage the structures of the network's inputs and hidden states. As a result, these  methods are less effective than recent methods that leverage the structures, such as SpatialDropout~\cite{spatial_dropout} and DropBlock~\cite{drop_block}, which randomly drop the values at certain contiguous areas in the hidden states and setting them to zero. Although the locations of dropout areas are random, the patterns of SpatialDropout and DropBlock are manually designed and fixed. Here we propose \emph{AutoDropout}, which  automates the process of designing dropout patterns. In our method, a controller learns to generate a dropout pattern at every channel and layer of a target network, such as a ConvNet or a Transformer. The target network is then trained with the dropout pattern, and its resulting validation performance is used as a signal for the controller to learn from. We show that this method works well for both image recognition on CIFAR-10 and ImageNet, as well as language modeling on Penn Treebank and WikiText-2. The learned dropout patterns also transfers to different tasks and datasets, such as from language model on Penn Treebank to Engligh-French translation on WMT 2014. Our code will be available.\footnote{Code repository: \url{https://github.com/google-research/google-research/tree/master/auto_dropout}.}
\end{abstract}

\section{\label{sec:intro}Introduction}
Modern neural networks are often over-parameterized \citep{deep_double_descent} and thus require proper regularization to avoid overfitting. A common regularization method is Dropout~\citep{dropout}, which randomly selects neurons from some intermediate layers of a network and replaces the values of these neurons with zero. In other words, we drop these neurons out of the current step of training. More recent studies show that imposing certain structures to the dropped neurons can lead to significant improvements over dropout neurons uniformly at random \citep{stochastic_depth,spatial_dropout,drop_block,variational_dropout_rnn,nas_module,rnn_regularization,transformer}. In practice, however, the dropout patterns are adapted to become different for different applications.

%Notably, the good structures to impose on the neurons to drop differ from domain to another and from one model architecture to another.

For example, in the text domain, \citet{rnn_regularization} suggest that, for a multi-layered LSTM \citep{lstm}, it is better to %not drop the neurons along the temporal dimension.
only drop the neurons in the vertical connections %inter-layer connections, i.e., vertical dropout, 
than to drop the neurons everywhere. 
\citet{variational_dropout_rnn} later propose Variational Dropout, where they drop neurons everywhere in the network but share a dropout pattern along the temporal dimension.
Both methods, however, are not used in the recent Transformer architecture, which only uses vanilla Dropout.
The differences in how LSTM and Transformer implement Dropout suggest that dropout patterns need to be tailored to different model architectures in NLP.

In the image domain, vanilla Dropout is often only applied to the fully-connected layers within a ConvNet~\citep{res_net,wide_res_net,pyramid_net,se_net,inception}. % Meanwhile,
Other convolutional layers often require the dropout neurons to have particular structures. For example, stochastic depth \citep{stochastic_depth} drops the whole residual branch in residual networks, and DropPath \citep{nas_module} drops a whole branch in multi-branched convolutional cells. 
%By analyzing these dropout patterns, 
\citet{drop_block} propose DropBlock which drop contiguous squares of neurons in the convolutional layers. While DropBlock works well on ResNet-50 and AmoebaNet \citep{nas_reg_evolution}, it is not proven to be successful in more recent architectures such as EfficientNet~\citep{efficient_det} and EfficientDet~\cite{efficient_det,data_aug_object_detection}. Again, the differences in the way ConvNet architectures use dropout patterns suggest that they also need to be specialized to architectures.\looseness=-1

\begin{figure*}
\centering
\includegraphics[width=0.55\textwidth]{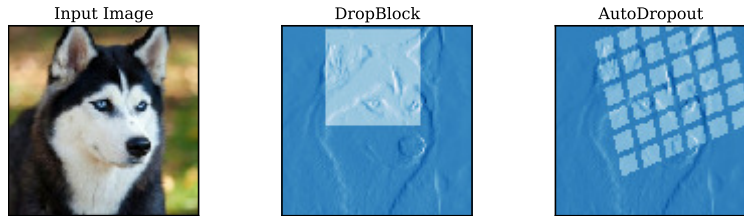}
\caption{\label{fig:example_drop}An example dropout pattern from our search space applied to an convolutional output channel. \textbf{Left:} the input image. \textbf{Middle:} DropBlock sets contiguous square blocks in the channel to zero. \textbf{Right:} a dropout pattern in the search space of \methodname. More patterns in our noise space are described in Section~\hyperref[sec:method]{Methods}.}
\end{figure*}

By studying the dropout patterns from previous works, we observe that these patterns are difficult to design and need to be specialized for each model architecture, task, and domain.
In this work, we address this difficulty by learning a specialized pattern for each model architecture, task, and domain.
To this end, we propose \emph{AutoDropout} which automates the process of designing specialized dropout patterns.
The main contribution of \methodname{} is a novel search space of structured dropout patterns.
In the search space we design, one can find a suitable for each model architecture and task.
Our search space generalizes many existing dropout patterns~\citep{dropout,variational_dropout,variational_dropout_rnn,stochastic_depth,drop_block}.
For example, Figure~\ref{fig:example_drop} shows a dropout pattern from our search space.
The pattern is generated by tiling a contiguous area and then transforming it geometrically.
The resulting pattern is applied to a convolutional output channel, which is a common building block of image recognition models.

Our implementation of \methodname{} has a controller that is trained by reinforcement learning (RL). The reward for the RL is the validation performance of the dropout pattern on a target network on a dataset of interest. We design a distributed RL-based search algorithm, which allows us to maximally leverage all machines available on an arbitrary cluster of computational nodes.\footnote{We will release the datasets consisting of the dropout patterns that our search algorithm has sampled and run. Like the similar datasets collected from benchmarking various model architectures \citep{nas_bench_101,nas_bench_201}.}

%Like other AutoML approaches \citep{nas,nas_evolution,nas_reg_evolution}, \methodname{} leverages intensive computational resources. Unlike the cases of searching for model architectures \citep{random_search_nas} and searching for data augmentation strategies \citep{rand_augment}, we find that random search \textit{cannot} find satisfying noise structures in our search space, hence necessitating a search algorithm. As such, we design a distributed RL-based search algorithm, which allows us to maximally leverage all machines available on an arbitrary cluster of computational nodes. As a result, our search procedure can operate up to 4,096 TPUv2 chips in parallel, utilizing more chips when they are available and fewer chips when they are busy.
% This allows us to massively parallelize the total search time of 37,000 TPU hours to finish in about 2 days. 37K=64*2*2*6*24, 4096=256*4*4

Our experiments show that \methodname{} can find dropout patterns that significantly improve commonly-used ConvNet and Transformer architectures.
On ImageNet, \methodname{} improves the top-1 accuracy of ResNet-50 from 76.5\% to 78.7\%, and EfficientNet-B7 from 84.1\% to 84.7\%. In the semi-supervised setting with CIFAR-10-4000, \methodname{} also improves the accuracy of Wide-ResNet-28-2 from 94.9\% to 95.8\%. For language modeling, \methodname{}  reduces the perplexity of Transformer-XL \citep{transformer_xl} on Penn Treebank from 56.0 to 54.9.

Additionally, when transferred to German-to-English translation on the IWSLT 14 dataset, the dropout pattern found by \methodname{} improves Transformer's BLEU score from 34.4 to 35.8, which is a new {\bf state-of-the-art on this dataset}. On English-to-French translation with WMT 2014, the transferred dropout pattern also yields an improvement of 1.9 BLEU scores over the Transformer model with vanilla Dropout. %To our knowledge, this is the first successful application of structured regularization 

Although the search cost of AutoDropout can be high, a simple use case of AutoDropout is to drop our found patterns into existing pipelines in the same way that AutoAugment policies~\cite{auto_augment} were used to improve state-of-the-art models.

\paragraph{Related works.} Our work has the same philosophy with existing neural architecture search and AutoAugment lines of research \citep{enas,darts,nas,nas_module,neural_optimizer_search,auto_augment,spec_augment,fast_auto_augment,efficient_net,nas_reg_evolution,device_placement,neural_combi,rand_augment,nas_evolution,proxyless_nas,automl_for_architecting}. We create a search space comprising the possible decisions and then use RL to search for the best decision.

More specifically, \methodname{} can also be viewed as data augmentation in the networks' hidden states. We generalize the successful approaches of searching for data augmentation \citep{spec_augment,auto_augment,rand_augment,fast_auto_augment} and apply them to the hidden states of ConvNets and Transformer networks. Unlike data augmentations, which are domain-specific, our dropout patterns for the hidden states have the same design philosophy on ConvNets for image recognition models and Transformer for text understanding models. CutMix \citep{cut_mix} and ManifoldMixup \citep{manifold_mixup} also apply successful data augmentation techniques such as CutOut \citep{cut_out} and Mixup \citep{mixup} into the hidden states. Implicit Semantic Data Augmentation (ISDA; \citet{isda}) approximate a Gaussian distribution of ConvNets' hidden states using the moving averages of their mean and standard deviations to generate more training examples.

\iffalse
In summary, our contributions are as follows:
\begin{enumerate}[leftmargin=0.04\textwidth]
\item We are the first to propose an automated search algorithm for dropout patterns.
\item We propose a general search space of dropout patterns to regularize neural networks. Our proposal works for both image recognition models and sequence models.
\item We propose a distributed RL-based search algorithm to learn the dropout pattern from our search space. We will release the datasets consisting of the noise structures that our search algorithm has sampled and run. Like the similar datasets collected from benchmarking various model architectures \citep{nas_bench_101,nas_bench_201}.
% \item We drastically improve the parallelism of distributed RL search. We also release the datasets consisting of the noise structures that our search algorithm has sampled and run. Like the similar datasets collected from benchmarking various model architectures \citep{nas_bench_101,nas_bench_201}, our datasets can serve further developments of searching for structured regularization noise. This is particularly important when our analysis suggests that random search does not yield satisfying results in our search space.
\item We achieve strong empirical results across many tasks, architectures, settings, and domains. Previously, only vanilla Dropout enjoys the same generality.
\end{enumerate}
\fi
\section{\label{sec:method}Methods}
\paragraph{Representing dropout patterns.} We represent the dropout patterns in our search space using elementwise multiplicative masks as adopted by many previous works \citep{dropout,variational_dropout,variational_dropout_rnn,stochastic_depth,nas_module,drop_block,transformer}. To bridge  the gap between training, when the mask is used, and inference, when the mask is not used, we scale the values of the non-drop neurons properly during training. Specifically, to apply a dropout pattern to a layer $h$ of a neural network, we randomly generate a binary mask $m$ of the same shape with $h$. We then scale the values in the mask $m$, and replace $h$ with:
\begin{equation}
\label{eqn:apply_noise}
\begin{aligned}
  \text{Drop}(h, m) \defeq h \otimes \left( \frac{\text{Size}(m)}{\text{Sum}(m)} \cdot m \right)
\end{aligned}
\end{equation}

\paragraph{Dimensional notations.} In modern deep learning frameworks \citep{tensorflow,pytorch}, intermediate layers are represented as high dimensional tensors. We denote the general shape of a tensor as $(N, d_1, d_2, ..., d_k, C)$, where $N$ is the \textit{batch} dimension, $C$ is the \textit{feature} dimension, and $d_1$, $d_2$, ..., $d_k$ are the \textit{spatiotemporal} dimensions. For instance, a layer in a typical ConvNet has a shape of $(N, H, W, C)$ where $H$ and $W$ are the layer's height and width; while a Transformer layer has the output of shape $(N, T, C)$ where $T$ is the temporal dimension which represents the number of tokens.%, i.e. the temporal dimension.

% Intuitively, the batch dimension $N$ and the feature dimension $C$ are different from the spatiotemporal dimensions $d_1$, $d_2$, ..., $d_k$. Specifically, the batch dimension $N$ controls the extent of data parallelism while training a network, and the feature dimension $C$ determines how much information should a particular layer in the network hold. Meanwhile, the spatiotemporal dimensions $d_i$'s depend on the network's input and can be mutually correlated. Thus, in our search space, we ignore the batch dimension $N$ and treat the dimension $C$ differently from the spatiodimensions $d_i$'s.

Our method is general and works well for both ConvNets and Transformers where the spatiotemporal dimensions are different from each other. In the following, we will first discuss the search space for ConvNets, and then discuss how we generalize it to Transformers.

\subsection{\label{sec:image_search_space}Search Space for Dropout Patterns in ConvNets}
\begin{figure}
\centering
\includegraphics[width=0.7\linewidth]{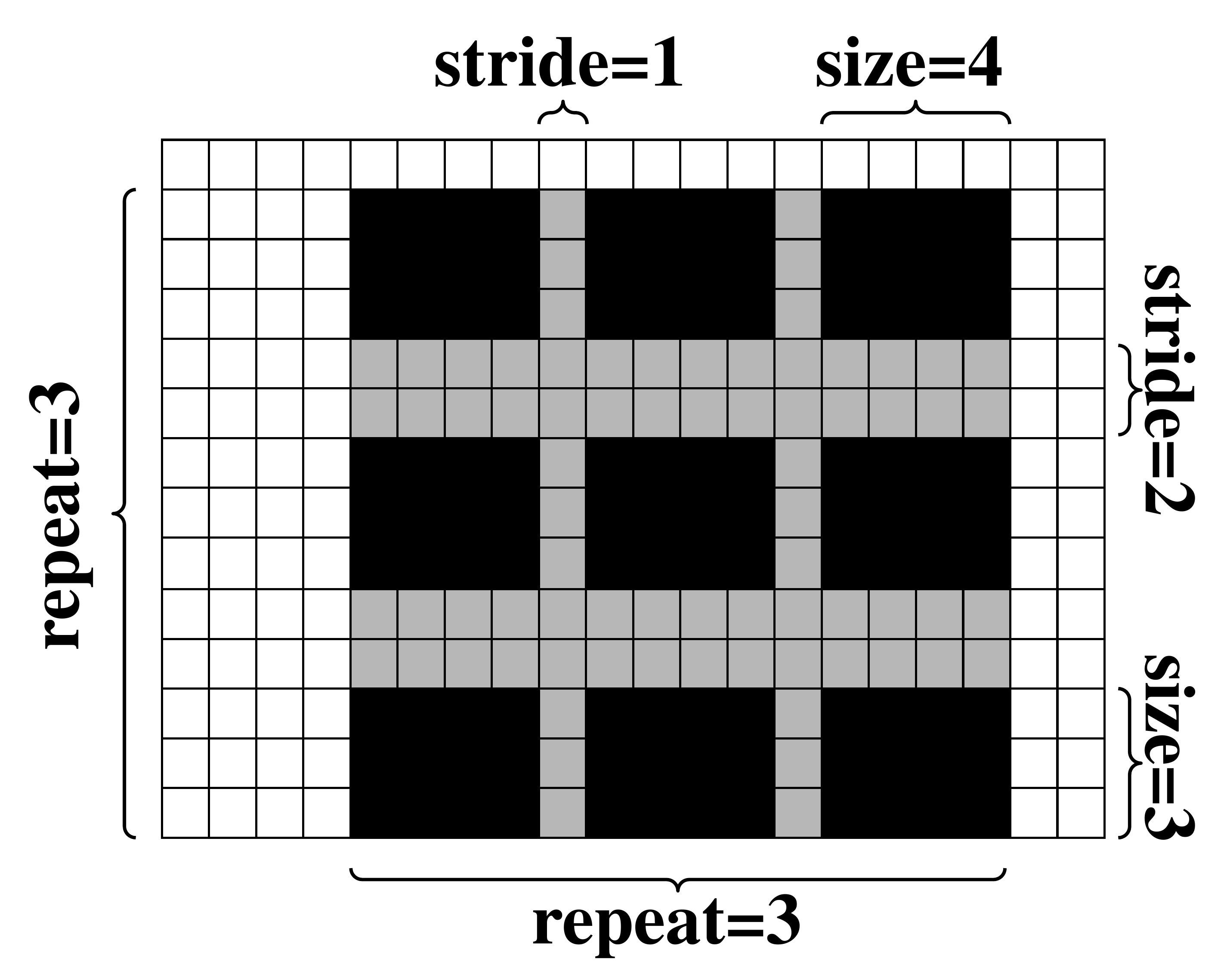}
\caption{\label{fig:tile_rectangles}Example of the basic patterns in our search space. A dropout pattern, represented in \textbf{black} and \textcolor{gray}{\textbf{gray}}, is applied to a grid of while cells representing the tensors. The neuron corresponding to the \textcolor{gray}{\textbf{gray}} cells are retained.}
\end{figure}
\paragraph{Basic patterns.} The basic pattern in our search space is a contiguous rectangle. The rectangle is then tiled to produce a dropout pattern. For ConvNets, the hyper-parameters that define the basic rectangle are two sizes height and width. The hyper-parameters that define the tiling are the stride and the number of repeats. Figure~\ref{fig:tile_rectangles} shows an example. For $C$ channels, we can either sample $C$ independent dropout patterns, or we can sample only one dropout pattern and then \textbf{share} it along the feature dimension.

\paragraph{Geometric transformations.} In addition to tiling the rectangles, we introduce two geometric transformations into our search space: rotating about the spatial center, and shearing along each spatial dimension. When the transformations result in fractional coordinates, we round them to the nearest integers.

\paragraph{Where to apply the dropout pattern.} Once we have a dropout pattern, there is a decision about where we should apply it to. Here, we apply the dropout pattern to the output of batch normalization layers because we empirically observe that applying the pattern elsewhere in the network often leads to unstable training during our search process. If there is a \textbf{residual} connection in the ConvNet to regularize, then there is a choice of whether we should apply the dropout pattern to the residual branch as well. We leave this decision to the controller. 
%In ConvNets with residual connections \citep{res_net,wide_res_net,efficient_net}, we additionally let our controller decide whether to apply the dropout pattern to the residual branch as well.
Appendix~\hyperref[sec:convnet_space]{Details on the Search Spaces for ConvNets} visualizes where the noise masks are applied in some network architectures in our experiments in Figure~\ref{fig:noise_where_image}, and specifies more details about our ConvNet search space.

% If a particular cell is repeated multiple times in the entire network, then the dropout pattern only depends on the spatial dimensions of the immediate tensors. For example, between stride convolutions and other dimensional reduction operations, the noise pattern are kept the same. Additionally, at dimensional reduction layers, if batch normalization is applied to the residual path to correct the resulting dimensions, we also apply the noise pattern after batch normalization layer.

\subsection{\label{sec:search_model}Controller Model and Search Algorithms}
\begin{figure*}[htb!]
\centering
\includegraphics[width=0.9\textwidth]{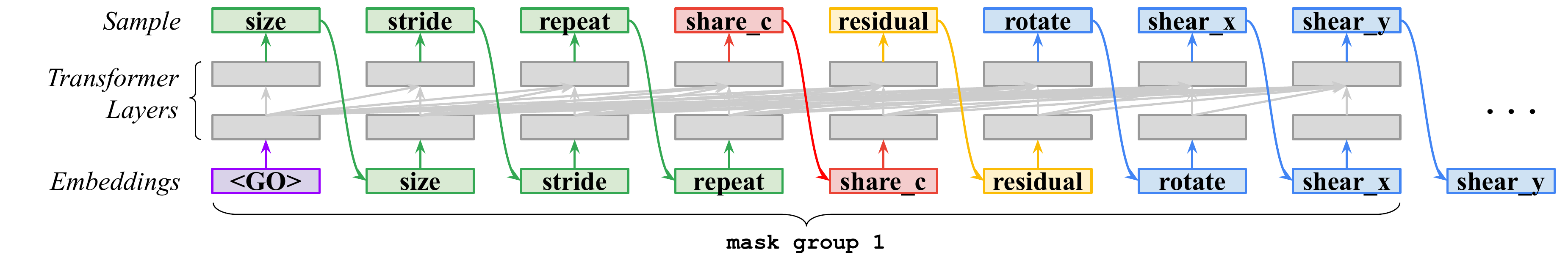}
\caption{\label{fig:controller_architecture}Our controller is a Transformer network. The network generates the tokens to describe the configurations of the dropout pattern. The tokens are generated like words in a language model. For every layer in a ConvNet, a group of 8 tokens need to be made to create a dropout pattern. These 8 tokens are generated sequentially. In the figure above, \texttt{size}, \texttt{stride}, and \texttt{repeat} indicate the size and the tiling of the pattern; \texttt{rotate}, \texttt{shear\_x}, and \texttt{shear\_y} specify the geometric transformations of the pattern; \texttt{share\_c} is a binary deciding whether a pattern is applied to all $C$ channels; and \texttt{residual} is a binary deciding whether the pattern is applied to the residual branch as well. If we need $L$ dropout patterns, the controller will generate $8L$ decisions.}
\end{figure*}
\paragraph{Model Architecture and Search Algorithm.} We parameterize our controller with a Transformer network, as illustrated in Figure~\ref{fig:controller_architecture}. We train the parameters $\theta$ of our controller using the REINFORCE algorithm with a standard moving average baseline \citep{reinforce}. That is, we optimize $\theta$ to minimize the objective via Monte Carlo gradient estimation:
\begin{equation}
\begin{aligned}
  J(\theta)
    &= \E_{r \sim P(\rvr; \theta)} [\text{Perf}(r)] \\
  \widehat{\nabla}_\theta J
    &= \frac{1}{M} \sum_{i=1}^{M} (\text{Perf}(r_i) - b) \cdot \nabla_\theta \log{P(r_i; \theta)}
\end{aligned}
\end{equation}
Here, $b$ is the moving average baseline, $M$ is the empirical batch size which we set to 16, and $\text{Perf}(r)$ is measured by training a target network with a dropout pattern $r$ on a designated proxy task's validation set. We find it important to tailor the proxy task according to the actual downstream task. We discuss our proxy tasks in detailed in~\hyperref[sec:exp]{Experiments}.

\paragraph{Improving Parallelism.} Previous works in architecture search and data augmentation search \citep{nas,nas_module,neural_optimizer_search,auto_augment,spec_augment,efficient_net} typically wait for minibatches of $M$ dropout patterns to finish training before making every update on $\theta$. Since each child model can take significantly long to train, and is subjected to multiple failures, such as jobs being descheduled on a shared cluster, waiting for $M$ dropout patterns to finish can cause an unnecessary bottleneck.

To alleviate this bottleneck, we propose a modification. Specifically, in a shared environment, the number of available machines will vary over time. Sometimes, the number of machines will be lower than $M$. In this case, we will have to use this low number of machines to slowly compute the rewards for $M$ configurations. However, sometimes the number of machines will be much higher than $M$. In such case, we want to generate many more than $M$ jobs to take advantage of the available resources. But even in such case, for training stability, we only use a minibatch of $M$ configurations, causing the other trained configurations to have stale gradient. To adjust for the staleness of their gradients, we need to reweigh the gradient properly as explained later. 

Our implementation maintains two queues:
%a modification. Specifically, our controller is given a maximum capacity $C$ of parallel child configurations to execute at any given time. The controller maintains two queues:
a queue $q_\text{unfinished}$ of unfinished jobs and a queue $q_\text{finished}$ of finished jobs. Whenever the $q_\text{unfinished}$ contains less than its capacity $C$, the controller generates $n = C - \abs{q_\text{unfinished}}$ new dropout patterns $r_1$, $r_2$, ..., $r_n$ and fills up $q_\text{unfinished}$ with the pairs $(r_i, P(r_i; \theta_i))$, where $\theta_i$ is the value of the controller's parameters at the time $r_i$ is sampled.

On the other hand, whenever a dropout pattern $r$ finishes training, the controller dequeues $(r, \text{Perf}(r))$ from $q_\text{unfinished}$ and moves it into $q_\text{finished}$. Whenever the capacity $\abs{q_\text{finished}}$ reaches $M$, $M$ configurations along with their accuracy are dequeued from $q_\text{finished}$ to perform an update on $\theta$. The caveat of this approach is that due to many dropout patterns being executed in parallel, the controller parameter $\theta$ when we update the controller with a configuration $r_i$ can be different from the $\theta_i$ when $r_i$ was generated. To account for this difference, we resort to importance sampling, which allows us to write the training objective $J(\theta)$ as follows:
\begin{equation}
\label{eqn:controller_queue_training}
\begin{aligned}
  &\nabla_\theta J(\theta)
    = \nabla_\theta \E_{r \sim P(\rvr; \theta)} [\text{Perf}(r)] \\
    &\approx \frac{1}{M} \sum_{i=1}^{M} \text{Perf}(r_i) \cdot \frac{P(r_i; \theta)}{P(r_i; \theta_i)} \cdot \nabla_\theta \log{P(r_i; \theta)}
\end{aligned}
\end{equation}
Implementing this update rule simply requires scaling the gradient $\nabla_\theta \log{P(r_i; \theta)}$ by the ratio of the two probabilities as shown in Equation~\ref{eqn:controller_queue_training}. In our design, the only training bottleneck is the number of workers that can be run in parallel. In practice, distributed search procedures like ours typically run on a shared cluster, where the number of available workers varies instantly. Our design obviates the need to reserve all $C$ workers throughout the search procedure and allows us to use a large value of $C$ to achieve better parallelism when more workers are available.

\subsection{\label{sec:text_search_space}Search Space for Dropout Patterns in Transformers}
% Compared to image recognition models, text processing models have a more diverse design choice. Text data can be processed by recurrent network \citep{seq2seq,rnn_regularization}, convolutional networks \citep{conv_seq2seq,conv_nmt}, or by attention-based networks \citep{transformer,transformer_xl}. In this paper, we focus on the Transformer architecture as it is commonly used across many tasks \citep{transformer,bert,transformer_xl}.

\begin{figure*}[htb!]
\centering
\includegraphics[width=0.7\textwidth]{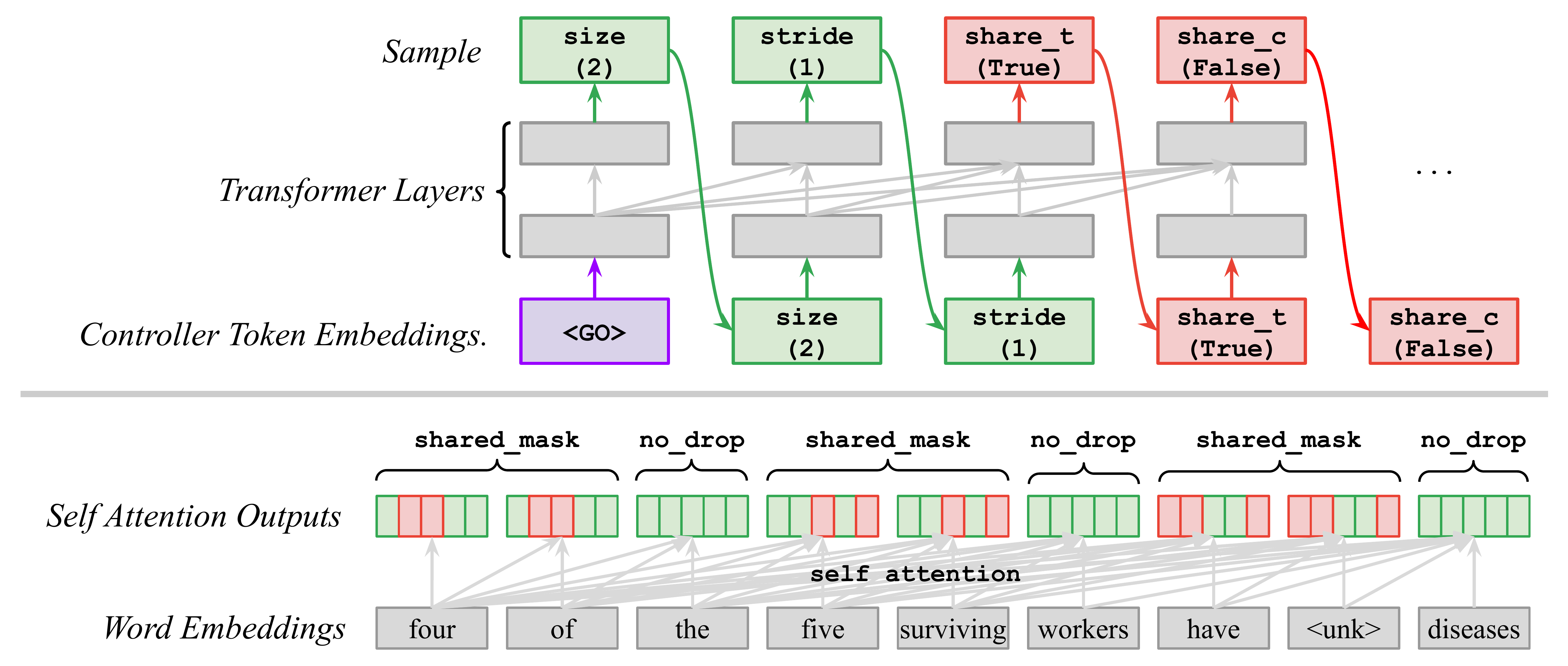}
\caption{\label{fig:text_example}An example of our controller generating ad dropout pattern for a self-attention operation. \textbf{Top:} the controller's outputs. The tokens have the following meanings: \texttt{size=2} and \texttt{stride1} means that the dropout pattern affects two consecutive tokens, then skips two token, then affects the next two consecutive tokens, and so on; \texttt{share\_t=True} means that every block of two consecutive tokens that the dropout pattern affects shares the same dropout mask; \texttt{share\_c=False} means that each of the $C$ feature dimensions of the $(N, T, C)$ tensor has its own independent mask.
% Similar to the dropout pattern of ConvNets, \texttt{size}, \textbf{stride}, and \textbf{repeat} indicate the size and the tiling of the pattern; \textbf{share\_c} is a binary deciding whether a pattern is applied to all $C$ channels.
\textbf{Bottom:} The dropout pattern that the controller's outputs realize on the self-attention operation. The values in the \textcolor{BrickRed}{\textbf{red}} cells are set to zero, while the values in the \textcolor{ForestGreen}{\textbf{green}} are kept intact.}
\end{figure*}

\paragraph{Basic patterns.} Intermediate layers in Transformer models typically have three dimensions $(N, T, C)$, where $N$ and $C$ are the batch dimension and the channel dimension, similar to those of ConvNets, and $T$ is the number of tokens, such as words or sub-word units.
The dropout pattern for this dimension $T$ is realized by generating four hyper-parameters: \texttt{size}, \texttt{stride}, \texttt{share\_t}, and \texttt{share\_c}. \texttt{size} indicates how many tokens does a pattern affects; \texttt{stride} indicates the number of tokens to be skipped by the pattern; \texttt{share\_t} is a binary deciding whether all the tokens covered by \texttt{size} are set to zero using the same noise mask or independent noise masks; and \texttt{share\_c} is a binary deciding whether a the dropout pattern shared along the channel dimension $C$.
% Despite its simplicity,
Once the values of \texttt{size}, \texttt{stride}, \texttt{share\_t}, and \texttt{share\_c} are decided, at each training step, we sample the starting position to apply the resulting dropout pattern. We repeat the pattern until the end of the sequence, following \texttt{size} and \texttt{stride}. 
Figure~\ref{fig:text_example} provides an illustration of a dropout pattern that our controller samples from our search space, and how the pattern is applied to a sequence of words.

Many successful regularization patterns for text processing models are included in our basic patterns. For instance, WordDropout \citep{word_dropout} can be realized from our patterns by setting \texttt{share\_c=True}, while Variational Dropout \citep{variational_dropout_rnn} can be realized by setting \texttt{share\_t=True} and setting $size$ to the $T$, number of tokens in the sequence.

\paragraph{Where to apply the dropout pattern.} Unlike the case for image recognition models, we find that the dropout patterns in our search space can be flexibly applied at multiple sub-layers within a Transformer layer (e.g., on the query, key, value, softmax, output projection, and residual). As a result, we apply one independent dropout pattern to each of them. 
%To allow the controller to \textit{not} apply the noise pattern everywhere, we add a special choice of repeating the rectangular blocks for $0$ times. 
Figure~\ref{fig:noise_where_text} in our Appendix~\hyperref[sec:text_search_space_illustration]{Details on the Search Spaces for Transformer} specifies all the possible places to apply the dropout patterns in our Transformer model.
% In total, for a Transformer network, the controller needs to generate 52 decisions per layer (4 tokens $\times$ 13 decisions per layer).
We will use this pattern at all Transformer layers in the Transformer network. In our implementation, \texttt{size} is overloaded, and if it has the value of zero, the dropout pattern is not applied at the corresponding.
\section{\label{sec:exp}Experiments}
In the following sections, we will apply \methodname{} to both ConvNets and Transformers. For ConvNets, we first consider \hyperref[sec:image_sup]{Supervised Image Classification} and then we consider \hyperref[sec:image_semi_sup]{Semi-supervised Image Classification}. For Transformer, we consider \hyperref[sec:language_model]{Language Model and Machine Translation applications}. Finally, we compare our search method against random search.

\subsection{\label{sec:image_sup}Supervised Image Classification with ConvNets}
We first evaluate \methodname{} on two standard benchmarks for image classification: CIFAR-10 \citep{cifar10} and ImageNet \citep{imagenet}. For CIFAR-10, we use Wide ResNet 28-10 (WRN-28-10; \citet{wide_res_net}) because it is a common baseline on this dataset. For ImageNet, we consider ResNet-50 \citep{res_net} because it is a common architecture for ImageNet. We also consider EfficientNet \citep{efficient_net} since it is closer to the state-of-the-art than ResNet. For each benchmark and model, we first use \methodname{} to search for a good dropout pattern on a proxy task, and then scale up the best found pattern to apply to the final task.

\paragraph{Search Configurations and Proxy Tasks.} We refer readers to Appendix \hyperref[sec:hparams]{Hyper-parameters of Experiments} for detailed training information of our controller. Here, we focus on the proxy tasks that we design to reduce the \methodname's search time. We scale down the final architecture and reduce the amount of data for the final task as follows.\looseness=-1

For CIFAR-10, we search with a WRN-28-2 on the entire dataset, reserving 10\% of the original training set for validation. For ImageNet, we scale down ResNet-50 and EfficientNet-B0 so that each of their layers has half the number of channels as the original models. We use 80,000 examples for training and 5,000 examples for validation. The controller's reward is the accuracy of the dropout pattern on the validation set. We train each dropout pattern on CIFAR-10 for 32,000 steps, and train each pattern on ImageNet for 16,000 steps. Under these settings, each dropout pattern trains in approximately 40 minutes on both datasets. Our search explores 16,384 patterns for each task. % We use the default image size of 32x32 for CIFAR-10 and 224x224 for ImageNet.

\paragraph{Baselines.} For WRN-28-10 and ResNet-50, we compare \methodname{} against DropBlock \citep{drop_block}, since DropBlock has been well-tuned for these models. For EfficientNet, we compare \methodname{} with Stochastic Depth \citep{stochastic_depth} since it is the default noise-based regularization scheme of this architecture. We implement these baselines in our environment for fair comparison. Note that large EfficientNet models, such as B3, B5, B7 in our experiments, enlarge the spatial dimensions of the input images. For these models, we proportionally scale up the sizes and strides of the masks found by \methodname{} on these models. Training details of all models are in our Appendix~\hyperref[sec:hparams]{Hyper-parameters of Experiments}.

\begin{table*}[htb!]
\small
\centering
\resizebox{0.9\linewidth}{!}{%
  % \begin{tabular}{lllll}
  \begin{tabular}{lccccccc}
  \toprule
  \multirow{2}{*}{\vtop{\hbox{\textbf{Regularization}}\vspace{0.01\linewidth}\hbox{\textbf{Methods}}}} &
  \textbf{CIFAR-10} &
  % \textbf{CIFAR-100} &
  \multicolumn{5}{c}{\textbf{ImageNet}} \\
  \cmidrule(lr){2-2} \cmidrule(lr){3-7}
  &
  WRN-28-10 &
  ResNet-50 &
  EfficientNet-B0 &
  EfficientNet-B3 &
  EfficientNet-B5 &
  EfficientNet-B7\\
  \midrule
  None &
  96.1 $\pm$ 0.12 &
  76.5 / 93.4 & 76.2 / 92.9 & $-$ & $-$ & $-$ \\
  DropBlock \citep{drop_block} &
  96.2 $\pm$ 0.07 &
  78.3 / 94.3 & 
  76.3 / 92.8 &
  $-$ &
  $-$ &
  $-$ \\
  Stochastic Depth \citep{stochastic_depth} &
  96.2 $\pm$ 0.07 &
  77.5 / 93.7 &
  76.8 / 93.1 & 
  80.2 / 95.0 &
  82.5 / 96.2 &
  84.1 / 96.9 \\
  \methodname{} &
  \textbf{96.8 $\pm$ 0.09} &
  \textbf{78.7 / 94.3} &
  \textbf{77.5 / 93.8} &
  \textbf{80.9 / 95.6} &
  \textbf{83.1 / 96.5} &
  \textbf{84.7 / 97.1} \\
  \bottomrule
  \end{tabular}
}%
\caption{\label{tab:image_sup}Performance of \methodname{} and the baselines on supervised image classification (higher is better). This is a control experiment and all models are implemented by us.}
\end{table*}

\paragraph{Results.} Figure~\ref{tab:image_sup} reports the results of our control experiments on ResNets and EfficientNet.  From Table~\ref{tab:image_sup}, it can be seen that \methodname{} outperforms DropBlock by 0.6\% accuracy on CIFAR-10 with WRN-28-10, which corresponds to a 16\% error reduction. Notably, on CIFAR-10 with WRN-28-10, DropBlock does not yield significant improvements compared to not using regularization at all, suggesting that the intuition on blocking contiguous regions is not sufficient. On ImageNet, \methodname{} improves the top-1 accuracy of ResNet-50 on ImageNet by 0.4\% compared to DropBlock. \methodname{} improves the accuracy of all EfficientNet models by a margin of 0.7\% on average. This is larger than the improvement of 0.5\% that DropBlock delivers on AmoebaNet \citep{drop_block,nas_reg_evolution}, even though EfficientNet baselines have higher accuracy than AmoebaNet.

\paragraph{Pushing the limits of ResNets.} In the above experiments, we wanted to perform fair comparisons against other baselines, and did not combine \methodname{} with any data augmentation methods. Here, we aim to push the limits of WRN-28-10 and ResNet-50 by combining \methodname{} and other data augmentation methods. As such, we apply the pattern found by \methodname{} on CIFAR-10 with RandAugment \citep{rand_augment} to WRN-28-10 and achieve 97.9\% accuracy. We also apply the pattern found by \methodname{} on ImageNet with RandAugment and achieve 80.3\% top-1 accuracy with ResNet-50 on ImageNet. These results outperform existing state-of-the-art results on these datasets with the same model architectures, as presented in Table~\ref{tab:image_sup_sota}.

Table~\ref{tab:image_sup_sota} also shows that \methodname{} is the only method that improves the performance on both CIFAR-10 with WRN-28-10 and ImageNet with ResNet-50. Among other baselines, Manifold Mixup \citep{manifold_mixup} improves the CIFAR-10 accuracy but has a weak accuracy on ImageNet. Meanwhile, CutMix \citep{cut_mix} achieves good accuracy on ImageNet but worsens CIFAR-10 accuracy. These observations suggest that regularization methods that are validated for a certain architecture and dataset might not deliver as strong performance for another architecture and dataset, necessitating automated designing procedures like \methodname.

\begin{table}[htb!]
\centering
\resizebox{\linewidth}{!}{%
  % \begin{tabular}{lllll}
  \begin{tabular}{lll}
  \toprule
  \multirow{2}{*}{\textbf{Methods}} &
  \textbf{CIFAR-10} &
  \textbf{ImageNet} \\
  & (WRN-28-10) & (ResNet-50) \\
  \midrule
  Stochastic Depth \shortcite{stochastic_depth} & 96.2 $\pm$ 0.07${}^\dagger$ & 77.5 / 93.7 \\
  DropPath \shortcite{fractal_net} & 95.4 & 77.1 / 93.5 \\
  Manifold Mixup \shortcite{manifold_mixup} & 97.5 $\pm$ 0.02 & 77.5 / 93.8 \\
  Mixup \shortcite{mixup} & 97.1 $\pm$ 0.08${}^\dagger$ & 77.9 / 93.9 \\
  CutMix \shortcite{cut_mix} & 96.7 $\pm$ 0.05${}^\dagger$ & 78.6 / 94.1 \\
  MoEx \shortcite{moex} & 96.7 $\pm$ 0.03 & 79.1 / 94.3 \\
  CutMix+RandAugment \shortcite{rand_augment} & 97.0 $\pm$ 0.06${}^\dagger$ & 78.3 / 94.2${}^\dagger$ \\
  CutMix+FixRes \shortcite{fix_res} & n/a & 79.8 / 94.9 \\
  \midrule
  \methodname+RandAugment
  & \textbf{97.9 $\pm$ 0.06} & \textbf{80.3 / 95.1} \\  
  \bottomrule
  \end{tabular}
}%
\captionof{table}{\label{tab:image_sup_sota}Performance of \methodname{} and representative baselines on supervised image classification (higher is better). \textit{$(\dagger)$ denotes our implementation. CutMix+FixRes is not applicable for CIFAR-10 since we keep the image resolution at 32x32 for CIFAR-10.}}
\end{table}

\begin{table}[htb!]
\centering
% \raisebox{0.4\baselineskip}{
\resizebox{0.92\linewidth}{!}{%
\begin{tabular}{lll}
  \toprule
  \multirow{2}{*}{\textbf{Methods}} & \textbf{CIFAR-10-4K} & \textbf{ImageNet-10\%} \\
  & (WRN-28-2) & (ResNet-50) \\
  \midrule
  LGA+VAT \shortcite{lga} & 87.9 $\pm$ 0.19 & $-$ \\
  ICT \shortcite{ict} & 92.7 $\pm$ 0.02 & $-$ \\
  MixMatch \shortcite{mixmatch} & 93.8 $\pm$ 0.06 & $-$ \\
  ReMixMatch \shortcite{remixmatch} & 94.9 $\pm$ 0.04 & $-$ \\
  % MixMatch+NS3L \shortcite{ns3l} & 93.1 $\pm$ 0.12 & $-$ \\
  % PCL \shortcite{pcl} & $-$ & $-$ ~~~~/ 86.2 \\
  LLP \shortcite{llp} & $-$ & $-$ ~~~~/ 88.5 \\
  SimCLR \shortcite{simclr} & $-$ & 69.3 / 89.0 \\
  FixMatch \shortcite{fix_match} & \textbf{95.7 $\pm$ 0.05} & 71.5 / 89.1 \\
  % MetaPseudoLabels \shortcite{meta_pseudo_labels} & \textbf{96.1 $\pm$ 0.07} & \textbf{73.9 / 91.9} \\ 
  \midrule
  UDA \shortcite{uda} & 94.9 $\pm$ 0.12 & 68.8 / 88.8 \\  
  UDA+\methodname{} & \textbf{95.8 $\pm$ 0.04} & \textbf{72.9 / 91.4} \\  
  \bottomrule
\end{tabular}
}%
% }%
\captionof{table}{\label{tab:image_semi_sup}Performance of \methodname{} and representative baselines on semi-supervised image classification (higher is better).}
\end{table}

\paragraph{Qualitative analysis of good dropout patterns.} \methodname{} finds several patterns that are unexpected. For example, the best noise pattern found for ResNet-50 on ImageNet, which is visualized in Figure~\ref{fig:resnet50_mask} in our Appendix~\hyperref[sec:noise_visualizaion]{Visualization of Good Dropout Patterns}, only injects noise into the first and the last bottleneck convolutional blocks. These two blocks also have different noise patterns. This behavior is different from DropBlock \citep{drop_block}, where a fixed and predefined mask of size 7x7 is applied at every layer. Additionally, rotation is applied in the first block, but not in the last block, suggesting that \methodname{} finds that rotational invariance should be enforced at the first block, where most low-level feature extracting happens, rather than in the last block, where most features have become more abstract. To validate the decisions of \methodname, we vary the locations where the dropout patterns are applied and observe about 1\% drop in top-1 accuracy, which is significant for ResNet-50 on ImageNet.\looseness=-1

\subsection{\label{sec:image_semi_sup}Semi-supervised Image Classification with ConvNets}
\paragraph{Experiment Settings.} We now consider two typical benchmarks for semi-supervised image classification: CIFAR-10 with 4,000 labeled examples and ImageNet with 10\% labeled examples. Since our search procedure of \methodname{} on ImageNet, as described in \hyperref[sec:image_sup]{the previous section}, uses a subset of images in ImageNet-10\%, we simply take the same dropout patterns found in that setting. We make sure that ImageNet-10\% contain the 80,000 images that we perform the search on. On CIFAR-10, we repeat our \methodname{} search with 3,600 training examples and 400 validation examples.

\paragraph{Baselines and Results.} We apply \methodname{} into Unsupervised Data Augmentation (UDA; \citet{uda}), since UDA has a simple implementation. As shown in Table~\ref{tab:image_semi_sup}, the dropout patterns found by \methodname{} improves UDA by 0.9\% on CIFAR-10 and 4.1\% Top-1 accuracy on ImageNet. Here we compare against recent representative strong baselines and skip earlier works such as~\cite{mean_teacher,vat,pseudo_label}.

\subsection{\label{sec:language_model}Language Model and Machine Translation}
%We have seen that the dropout pattern found in our search space can substantially boost the performance of many models under different settings. In this section, we show that \methodname{} can also find good dropout patterns to regularize text processing model models.

\begin{table*}[htb!]
\small
\centering
\resizebox{0.9\linewidth}{!}{%
  \begin{tabular}{lcccccc}
  \toprule
  \multirow{2}{*}{\textbf{Methods}} &
  \multicolumn{2}{c}{\textbf{PennTreebank}} &
  \multicolumn{2}{c}{\textbf{WikiText-2}} &
  \textbf{IWSLT-14 DeEn} &
  \textbf{WMT-14 EnFr} \\
  \cmidrule(lr){2-3}
  \cmidrule(lr){4-5}
  \cmidrule(lr){6-6}
  \cmidrule(lr){7-7}
  & ValidPPL (\textcolor{red}{$\downarrow$})&
  TestPPL (\textcolor{red}{$\downarrow$})&
  ValidPPL (\textcolor{red}{$\downarrow$})&
  TestPPL (\textcolor{red}{$\downarrow$})&
  BLEU (\textcolor{blue}{$\uparrow$}) &
  BLEU (\textcolor{blue}{$\uparrow$}) \\
  \midrule
  Transformer / Transformer-XL & 59.1 & 56.0 & 63.9 & 61.4 & 34.4 & 38.1 \\  
  +\methodname &
  \textbf{58.1} &
  \textbf{54.9} &
  \textbf{62.7} &
  \textbf{59.9} &
  \textbf{35.8} &
  \textbf{40.0} \\  
  \bottomrule
  \end{tabular}
}%
\caption{\label{tab:text_model}Performance of Transformer and Transformer-XL models trained with default regularization techniques vs. trained with \methodname{}. For PTB and WikiText-2, we report the model's perplexity (lower is better \textcolor{red}{$\downarrow$}). For IWSLT-14-DeEn and WMT-14-EnFr, we report BLEU scores (higher is better \textcolor{blue}{$\uparrow$}).}
\vspace{-0.015\textwidth}
\end{table*}

In this section, we first apply \methodname{} to regularize Transformer-XL \citep{transformer_xl} the task of language model on the Penn Treebank dataset (PTB; \citet{penntreebank}). PTB is a small dataset, with about 929K training tokens, 73K validation tokens, and 82K test tokens, on a vocabulary size of 10K. The small size of PTB makes it a  suitable testbed for \methodname.

\paragraph{Search Configuration.} We use the search space for Transformer models as described in the Section \hyperref[sec:text_search_space]{Search Space for Dropout Patterns inTransformers}. Every dropout pattern that our controller sampled is employed to regularize a training process of Transformer-XL \citep{transformer_xl}. We use the same model size as specified by \citet{transformer_xl}. We train every configuration from scratch for 160,000 steps, using a batch size of 16 and a segment length of 70. We use the cosine learning rate schedule so that each trial converges to a reasonable perplexity. On 4 TPU v2 chips, each of our runs takes about 40 minutes. The performance of a configuration $r$ is computed by $\text{Perf}(r) = 80 / \text{ValidPPL}(r)$.

\paragraph{Results.} Our results for Transformer models are reported in Table~\ref{tab:text_model}. Once again, hyper-parameters for each experiment are reported in our Appendix~\hyperref[sec:hparams]{Hyper-parameters of Experiments}. First, we take the dropout pattern that achieves the lowest perplexity on PTB and train for 300,000 steps. We compare our results with Variational Dropout, which is originally used by Transformer-XL \citep{transformer_xl}. Under this setting, \methodname{} outperforms Variational Dropout by 1.1 perplexity, which is a significant improvement on this dataset.

\paragraph{Transfer learning results.} To test the transferability of the found pattern, we also transfer it to three other tasks: 1) language modeling on WikiText-2 \citep{pointer_sentinel}, 2) German-English translation on the IWSLT-14 dataset, and 3) English-French translation on the WMT-14 dataset. On Wiki-Text-2, we compare \methodname's dropout pattern against Variational Dropout because we find that it works better than vanilla Dropout on this task. On translation tasks, we compare \methodname's dropout pattern against the vanilla Dropout configurations that are typically applied in Transformer models \citep{transformer}.

\paragraph{Qualitative analysis of the \methodname's dropout pattern.} For Transformer models, \methodname{} assigns different sizes and strides at different sub-layers in a Transformer layer. For instance, in our best dropout pattern, visualized in Figure~\ref{fig:transformer_mask} in our Appendix, \methodname{} learns that the pattern for the multi-head attention layer is similar to Variational Dropout \citep{variational_dropout_rnn}, but the pattern for the positional feed-forward layer follows word dropout \citep{word_dropout}. To validate that such decision is beneficial, we try to apply Variational Dropout in all layers of Transformer-XL and got the resulting validation perplexity of 59.8, which is 1.7 point higher than the configuration found by \methodname.

\subsection{\label{sec:random_search}Comparison with Random Search}
\begin{figure}[htb!]
\centering
\includegraphics[width=0.48\linewidth]{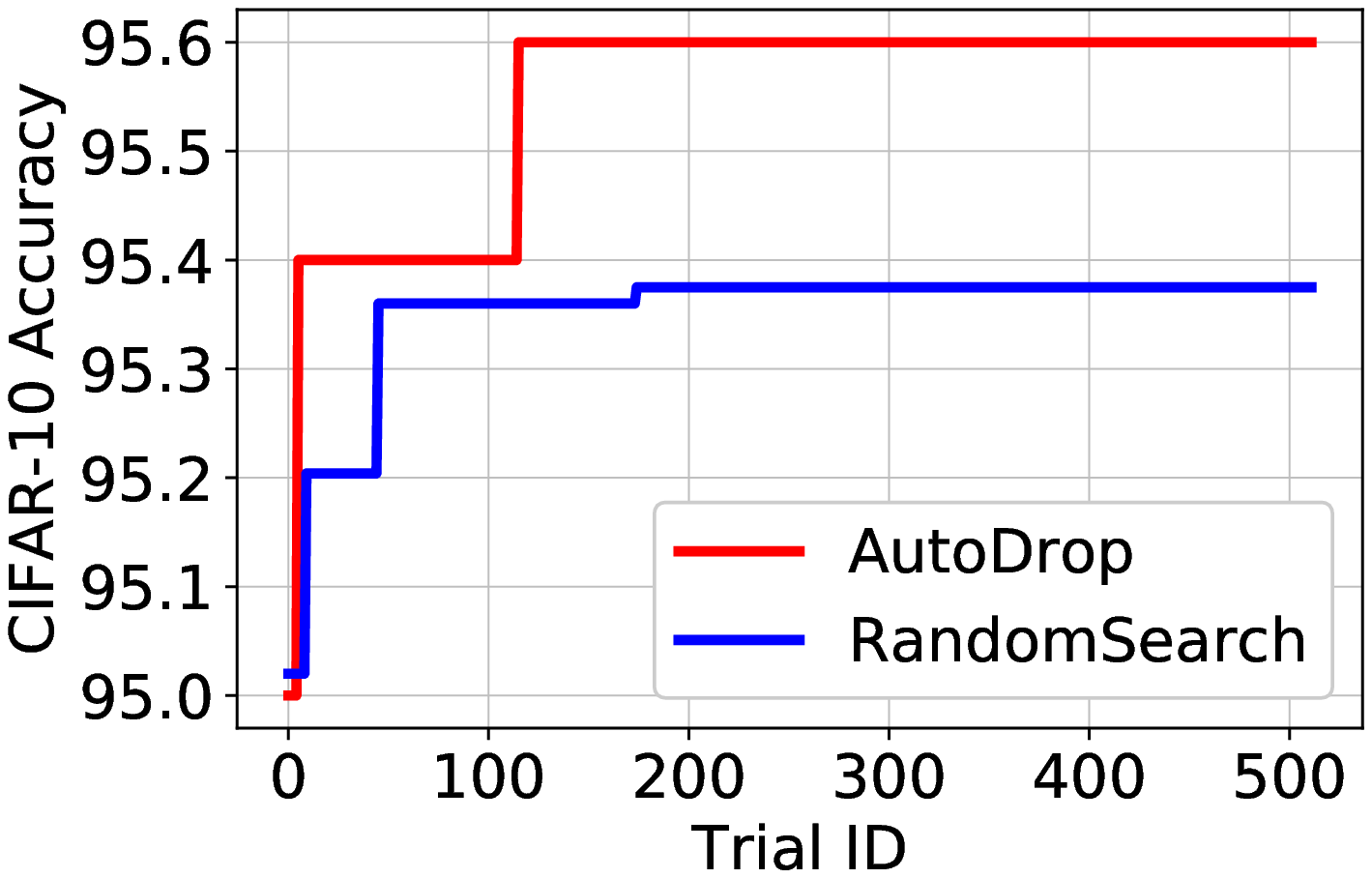}
\includegraphics[width=0.48\linewidth]{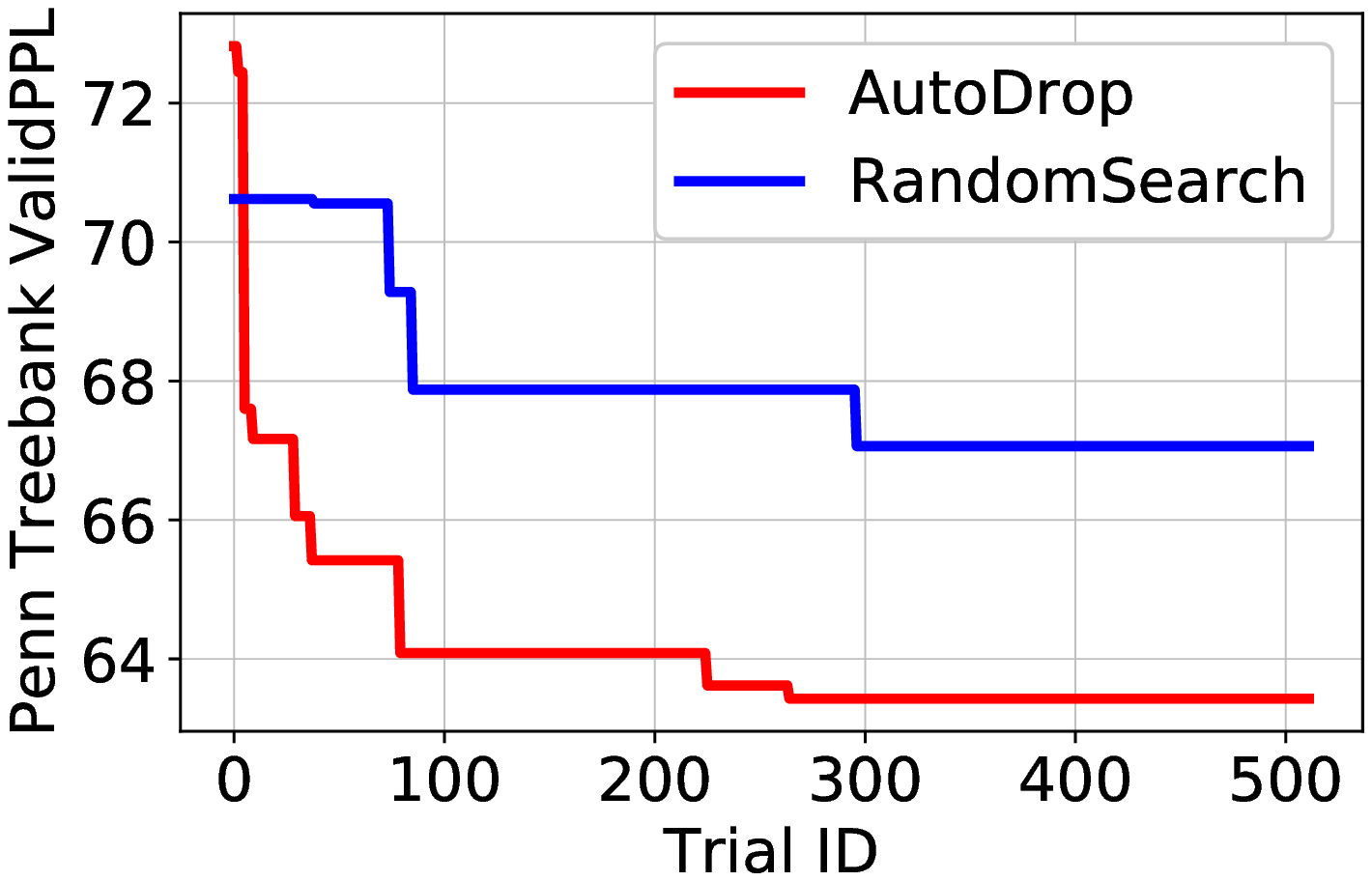}
\captionof{figure}{\label{fig:compare_with_rs}Best-so-far performances of the first 512 dropout patterns sampled by \methodname{} and by random search. \textbf{Top:} Accuracy on CIFAR-10 (higher is better); \textbf{Bottom:} ValidPPL on PennTreebank (lower is better).}
\end{figure}
Recent works on neural architecture search~\cite{random_search_nas} show that random search is a strong search baseline. Here we perform a controlled experiment to verify the advantage of \methodname's search process over random search. To this end, we sample 512 uniformly random patterns from the search space for WRN-28-2 on CIFAR-10 and another 512 uniformly random patterns from the search space for Transformer-XL on PTB. We train each of these patterns to convergence, and compare the results against training \textit{the first} 512 patterns suggested by \methodname{} under the same settings. In Figure~\ref{fig:compare_with_rs}, we plot the best-so-far performances of both methods, and observe substantial differences between \methodname{} and random search. Specifically, on CIFAR-10, the best patterns found by \methodname{} is more than 0.2\% accuracy above that of Random Search. Recall that from Table \ref{tab:image_sup}, we know that the standard deviation of CIFAR-10 accuracy in our code base is less than 0.1\%. This means that \methodname{} is more than 2x standard deviations away from random search and makes the difference significant. On PTB, the difference between \methodname{} and Random Search is more than 3 validation perplexity points, which is also significant for the dataset. We thus conclude that when searching for structured noise to regularize deep networks, RL search exhibits significant advantage compared to Random Search. %, and would be be eager to see progress on more efficient search approaches.
\section{\label{sec:conclusion}Conclusion and Future Directions}
We proposed \methodname, an algorithm to automatically design dropout patterns to regularize neural networks. Our algorithm successfully found dropout patterns that improve the performance of various ConvNets for image classification, as well as  Transformer models for language modeling and machine translation. Currently, a weakness of \methodname{} is that the method is computationally expensive. Therefore, a potential future direction is to develop more efficient search approaches, similar to the developments on architecture search \citep{enas,darts,proxyless_nas,pnas} and data augmentation search \citep{fast_auto_augment,isda,rand_augment}.

Although the search cost of \methodname{} can be high, a simple use case of AutoDropout is to reuse our found patterns in the same way that AutoAugment policies~\cite{auto_augment} were used to improve state-of-the-art models. To date, the method of reusing the found AutoAugment~\cite{auto_augment} and RandAugment~\cite{rand_augment} policies has benefitied many state-of-the-art models on CIFAR-10 and ImageNet (e.g.,~\citet{efficient_net,noisy_student,ridnik2020tresnet,foret2020sharpnessaware}).

\bibliography{main}

\appendix
\onecolumn
\begin{center}
  \LARGE{Appendix\\\textbf{\methodname: Learning Dropout Patterns to Regularize Deep Networks}}
\end{center}
\section{\label{sec:convnet_space}Details on the Search Spaces for ConvNets}
\paragraph{Details on generating dropout pattern.} In our search space for ConvNets, each dropout pattern is generated by its hyper-parameters: \texttt{size}, \texttt{stride}, \texttt{repeat}, \texttt{share\_c}, \texttt{residual}, \texttt{rotate}, \texttt{shear\_x}, and \texttt{shear\_y}, in that order. The available values for the operations are summarized in Table~\ref{tab:value_conv_net}.

\begin{table}[htb!]
\centering
\begin{tabular}{lll}
  \toprule
  \textbf{Parameters} &
  \textbf{Semantics} &
  \textbf{Available Values} \\
  \midrule
  \texttt{size} & Size of the pattern & $\{k \times \lfloor d / 5\rfloor: k = 0, 1, ..., 4\}$ ($d$ is the tensor's size) \\
  \texttt{stride} & How many cells to skip when tiling & $1, 2, 4, 8, 16$ \\
  \texttt{repeat} & How many times to repeat the pattern & $1, 2, 3, ..., 32$ \\
  \texttt{share\_c} & Share a pattern across channels & True, False \\
  \texttt{residual} & Apply a pattern to the residual branch & True, False \\
  \texttt{rotate} & Pattern's max rotating degree & $0, 15, 30, 45, 60, 75$ \\
  \texttt{shear\_x} & Pattern's max shearing rate & $0., 0.05, 0.1, ..., 0.55$ \\
  \texttt{shear\_y} & Pattern's max shearing rate & $0., 0.05, 0.1, ..., 0.55$ \\
  \bottomrule
\end{tabular}
\caption{\label{tab:value_conv_net}Semantics of the hyper-parameters that specify a ConvNet dropout pattern and their available values in our search space.}
\end{table}

We use the same dropout pattern for layers with same spatial size. For instance, in a ResNet-50, there are 4 bottleneck convolutions groups: having has 3 blocks, 4 blocks, 6 blocks, and 3 blocks respectively. The spatial dimensions of these blocks are 56x56, 28x28, 14x14, and 7x7, decreasing by a factor of 2 after each block due to strided convolutions or spatial reduction pooling. To reduce the number of decisions that our controller has to make, within each of these 4 groups, the dropout patterns are kept the same (but the actual samples of the dropout masks are still random at training time). Figure~\ref{fig:resnet50_mask} in~\hyperref[sec:noise_visualizaion]{Appendix} shows this sharing scheme in a pattern found by \methodname.

\begin{figure}[htb!]
\centering
\includegraphics[width=0.65\textwidth]{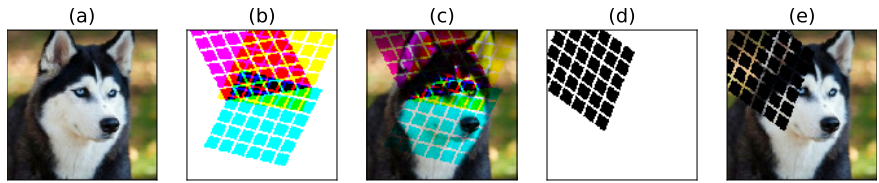}
\caption{\label{fig:image_example}Example noise patterns from our search space for image understanding models. Here, we visualize the noise applied on the RGB channels of an image for illustration purposes. \textbf{(a)}: The original image. \textbf{(b-c)}: Three different patterns, one for each of the RGB channels of an image, and the resulting image after applying the patterns. \textbf{(d-e)}: Three channels RGB share the same pattern (which leads to the color black), and the resulting image by applying this shared pattern. Image best viewed in color.}
\end{figure}
\paragraph{Example geometric transformations.} The geometric transformations, namely rotating and shearing along each dimensions, are implemented using projective transformations. Figure~\ref{fig:image_example} shows the effects of these transformations on some example dropout patterns. In this figure, we consider 3 RGB channels and visualize the patterns as they apply to the image. In our search space, the masks are applied to intermediate layers with many more channels.

\begin{figure}[htb!]
\centering
\includegraphics[width=0.9\textwidth]{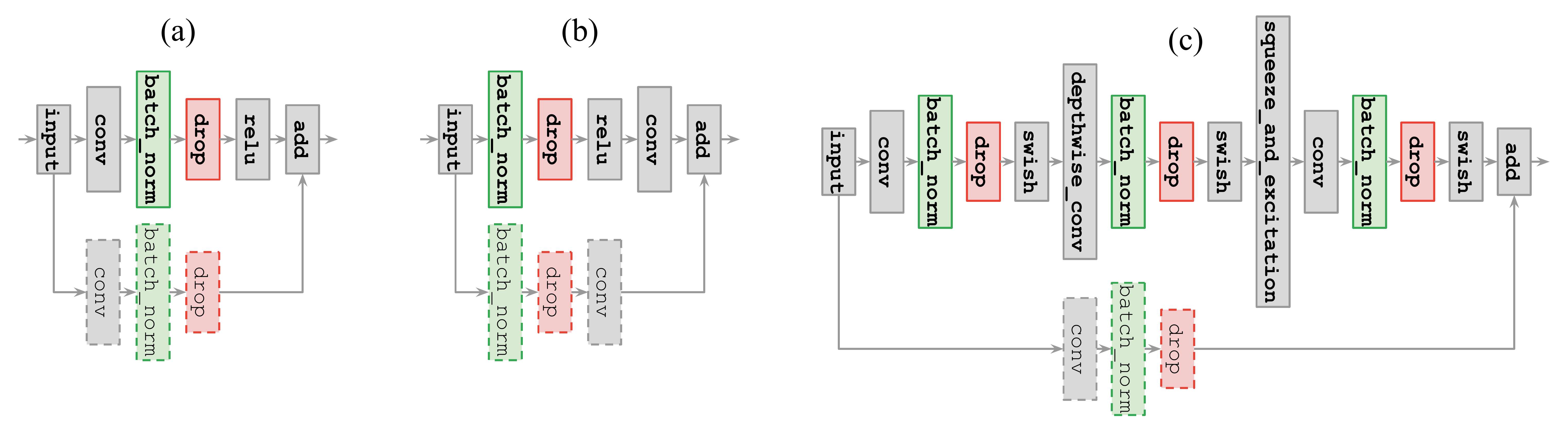}
\caption{\label{fig:noise_where_image}For ConvNets, we apply the dropout patterns immediate after the batch normalization layers. Shown are the examples ConvNet blocks in our experiments: \textbf{(a)} post-activation ResNet; \textbf{(b)} pre-activation ResNet; \textbf{(c)} Mobile Inverse Convolutional cell (MBConv; \citep{efficient_net}).}
\end{figure}
\paragraph{Where to apply the dropout patterns.} Figure~\ref{fig:noise_where_image} specifies where we apply the dropout patterns for ConvNets. In general, we apply the dropout pattern after each batch normalization layer. If a convolutional block has a residual branch, which sometimes has a 1x1 convolution followed by batch normalization, then we also apply a dropout pattern after the normalization as well.

\section{\label{sec:text_search_space_illustration}Details on the Search Spaces for Transformer}
\begin{table}[htb!]
\centering
\begin{tabular}{lll}
  \toprule
  \textbf{Parameters} &
  \textbf{Semantics} &
  \textbf{Available Values} \\
  \midrule
  \texttt{size} & How many consecutive tokens to affect & $0, 10, 20, 30, 40, 50, 60, 70$ \\
  \texttt{stride} & How many consecutive tokens to skip & $0, 5, 10, 15, 20$ \\
  \texttt{share\_t} & Share a mask across the tokens to affect & True, False \\
  \texttt{share\_c} & Share a pattern across channels & True, False \\
  \bottomrule
\end{tabular}
\caption{\label{tab:value_transformer}Semantics of the hyper-parameters that specify a Transformer dropout pattern and their available values in our search space.}
\end{table}
\paragraph{Details on generating dropout patterns.} In our search space for Transformer, each dropout pattern is generated by its hyper-parameters: \texttt{size}, \texttt{stride}, \texttt{share\_t}, and \texttt{share\_c}, in that order. The available values for the operations are summarized in Table~\ref{tab:value_transformer}.

\begin{figure}[htb!]
\centering
\includegraphics[width=0.7\textwidth]{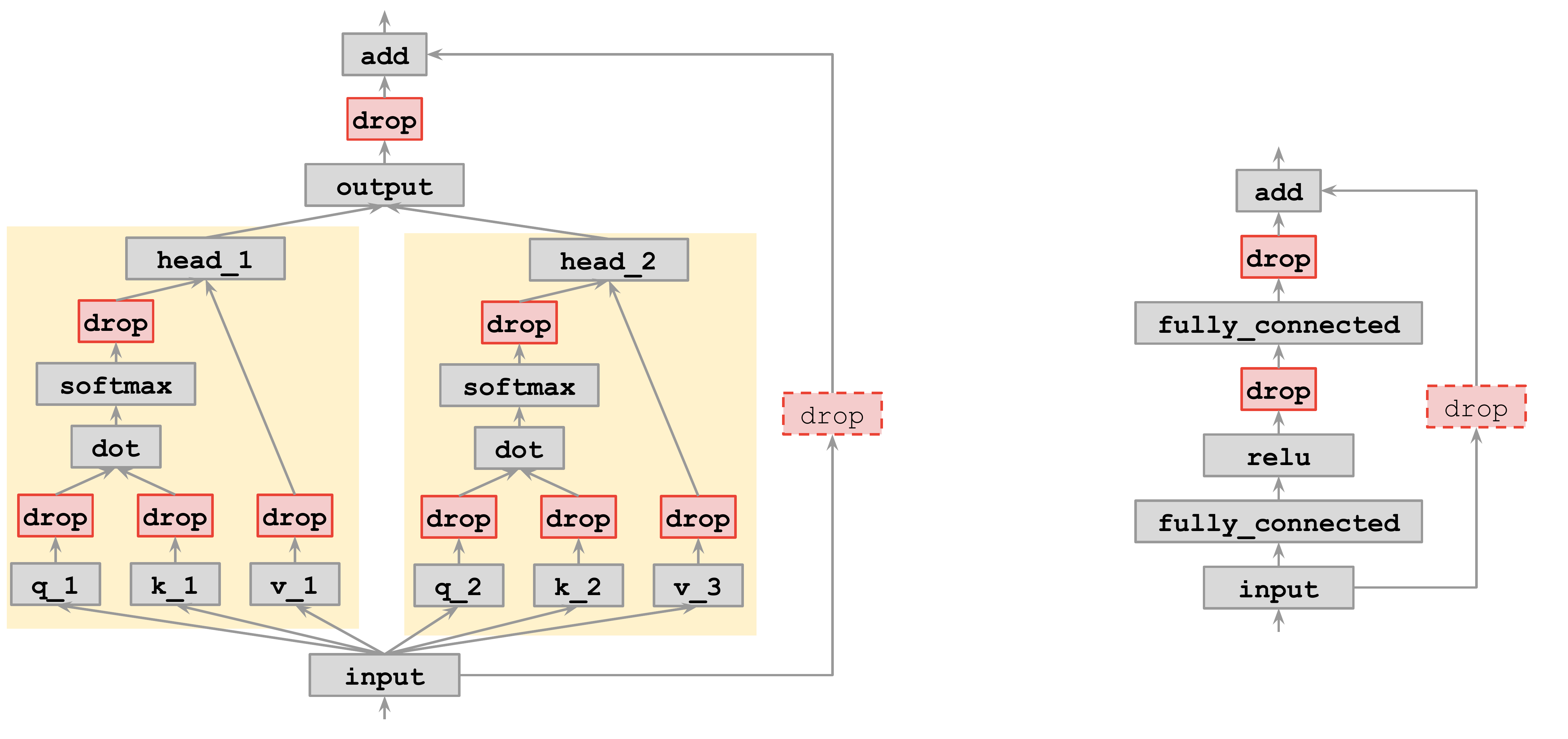}
\caption{\label{fig:noise_where_text}We apply the noise at various nodes inside a multi-head self-attention operation. \textbf{Left:} Noise in a two-headed attention operation. \textbf{Right:} Noise in a positional feed-forward operation.}
\end{figure}

We allow our controller to generate different patterns at different steps in a Transformer layer. Specifically, Figure~\ref{fig:noise_where_text} shows where the dropout patterns could be applied, in a self-attention operation and in a positional feed-forward operation. If a self-attention operation uses multi-head attention, then we use the same dropout pattern across all heads. However, within each head, the position to apply the dropout pattern is randomly sampled at training time. Similarly, in a typical Transformer network, where multiple Transformer layers are stacked above each other, we use the same dropout pattern to all layers, but the actual dropout mask are generated randomly and independently at each layer. 

\section{Illustration for the Controller Parallel Training Algorithm}
\begin{figure}[htb!]
  \centering
  \includegraphics[width=0.9\textwidth]{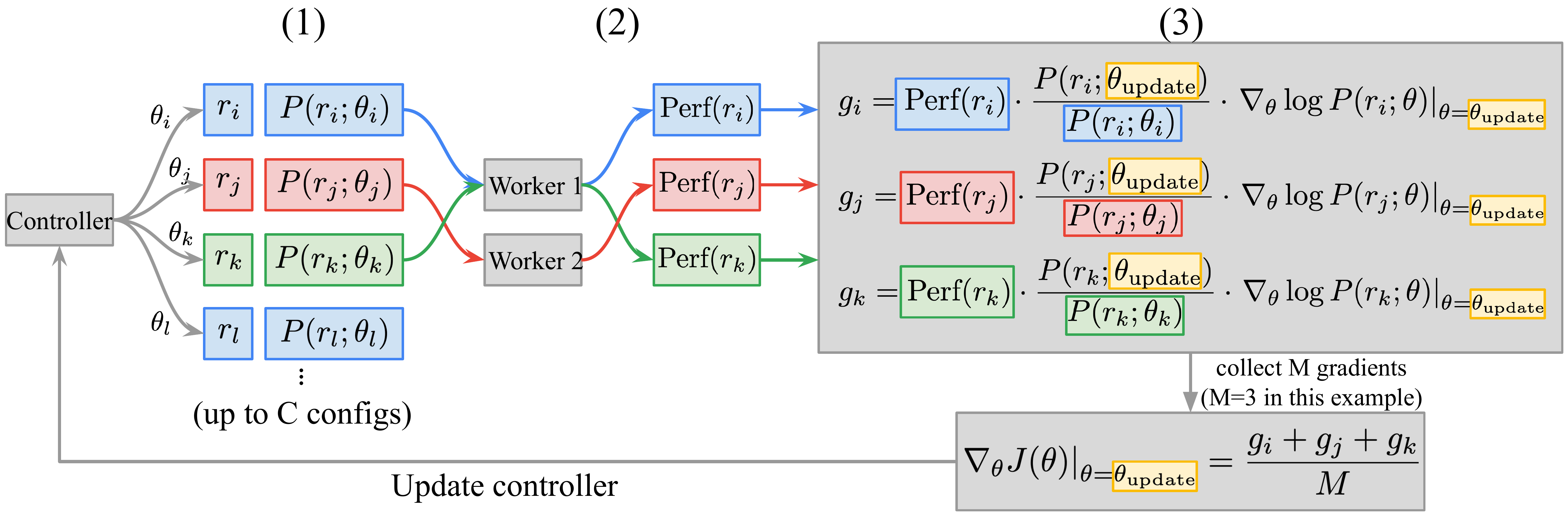}
  \caption{\label{fig:controller_queue}Illustration of our controller. From left tor right order with $M=3$ and $C \gg 3$. (1) The controller generates up to $C$ regularization rules and stores these rules in $q_\text{spawn}$, along with their sample probabilities. (2) A fixed pool of workers dequeue the rules from $q_\text{spawn}$ and train a model with these rules. In this case, $2$ workers dequeue and train $3$ rules $r_i$, $r_j$, $r_k$ to produce $\text{Perf}(r_i)$, $\text{Perf}(r_j)$, $\text{Perf}(r_k)$. (3) When $M$ rules have their $\text{Perf}$'s measured, their corresponding gradients are computed using importance sampling as in Equation~\ref{eqn:controller_queue_training}, and then are averaged to update the controller at its instantaneous parameter $\theta_\text{update}$. If we select $C$ sufficiently large, the only bottleneck of this procedure is the number of available workers.}
\end{figure}

We provide an illustrative working example of the distributed reinforcement learning search algorithm for controller, as described in Section~\hyperref[sec:search_model]{Controller Model and Search Algorithms}. Figure~\ref{fig:controller_queue} visualizes the example.

Our search process runs on a shared cluster. The controller has a queue $q_\text{unfinished}$ which stores the generated dropout patterns waiting to be executed, as the workers become available. In our example, suppose that thw cluster has only two available workers. The controller sequentially dequeues $(r_i, P(r_i, \theta_i))$ and $(r_j, P(r_j, \theta_j))$ and sends them to the two workers to train. When one of these dropout patterns finishes, say the $i^\text{th}$ on $\text{Worker}_1$ finishes, the controller sends $(r_i, \text{Perf}(r_i))$ into $q_\text{finished}$ and sends $(r_k, P(r_k, \theta_k))$ into the now-available $\text{Worker}_1$ to train. Later, after the $j^\text{th}$ dropout pattern and the $k^\text{th}$ dropout pattern both finish, and the controller has finished sending their results to $q_\text{finished}$, then $q_\text{finished}$ has $M=3$ finished configurations, where $M$ is the minibatch size for the controller updates. The controller now computes the gradients corresponding to the $i^\text{th}$, $j^\text{th}$, and $k^\text{th}$ dropout patterns, scales them according to the probability ratios as specified in Equation~\ref{eqn:controller_queue_training}, and averages the resulting gradients to update its parameters.

If during the controller's training, more workers become available, then more dropout configurations from $q_\text{unfinished}$ can be sent to the available workers to train to enhance the model's better parallelism. This is a significant advantage compared to previous AutoML search algorithms, which always requires $M$ workers to be available, or the search process has to stay idle waiting for the minibatches to finish.

\section{\label{sec:hparams}Hyper-parameters of Experiments}
\paragraph{Controller.} We use a small Transformer model to parameterize our controller. Specifically, our Transformer architecture has 4 layers, with the hidden size of 128. Each multi-head attention operation uses 4 heads, and each head has the hidden dimension size of 32. The positional feed-forward has the inner dimension of 32. The controller's parameters are initialized at a normal distribution with zero mean and a standard deviation of 0.02. We update the controller's parameters using Adam \citep{adam} with a constant learning rate of 0.00035 and the default values $\beta_1=0.9$, and $\beta_2=0.999$. We also use a moving average baseline with momentum $0.95$ to stabilize the update, and an entropy regularization of $10^{-5}$ to encourage the controller's explorations. For each search, our controller explores 16,384 dropout patterns in total, and updates its parameters using a batch size of 16, leading to 1,024 updates.

\paragraph{Image Recognition Models.} In order to avoid tuning the dropout rate at each layer of a ConvNet, we specify a single dropout rate \textit{for the final convolutional layer}. Previously layers have their dropout rate linearly increased from 0 to the specified value. During search time, we set the final value to 0.2. Once the search finishes, we tune the final value among the list of 0.1, 0.2, ..., 0.7. We find the with out dropout pattern, the ideal final dropout rate for WRN-28-10, ResNet-50, and EfficientNet are 0.6, 0.3, and 0.5. Apart from the layer-wise dropout rate, we use the same values with \citet{efficient_net} for EfficientNet, the same values with \citet{drop_block} for ResNet-50 on ImageNet, the same values with \citet{uda} for WRN-28-\{2,10\} on CIFAR-10. Note that this means that we train ResNet-50 for 240 epochs, which is 1.5 times longer than normally done for this architecture, but we train EfficientNet for 350 epochs, which is the same with \citet{efficient_net}.

\paragraph{Language Model.} For both Penn Treebank and WikiText-2, we use the Transformer-XL architecture \citep{transformer_xl}, which has 16 layers, hidden size of 380, 10 heads each of dimension 38, and positional feed-forward inner size of 900. For Penn Treebank, this results in a model with 24 million parameters, while for WikiText-2, this results in a model with 35 million parameters. We use a dropout rate of 0.5 for the embedding layer, a dropout rate of 0.6 for the softmax layer. We find these dropout rates from the Penn Treebank code released by \citet{transformer_xl}. We use the dropout rate of 0.2 elsewhere in our Transformer-XL model. We also use the state-value and state-difference regularizations \citep{awd_lstm}, even though we do not observe significant raise in perplexities without using them. We train with Adam for 160K steps during \methodname search, and 320K steps for the best architecture that \methodname{} finds. We using a cosine-decayed learning rate schedule \citep{cosine_lr}, starting at $3 \times 10^{-4}$ and decaying to $10^{-4}$ throughout 80\% of the training process. After the learning rate decays to $10^{-4}$, we continue the remaining 20\$ of the training process with a constant learning rate of $5 \times 10^{-5}$. During the last 20\% of the training procedure, we start collecting a moving average trail of the model's parameters. We perform one validation evaluation every 1,000 training steps and store the best model checkpoint. In the end, we obtain the test perplexity from the checkpoint with the lowest validation perplexity.

\paragraph{Machine Translation.} We use the Transformer-Base architecture from \citet{transformer}. We tokenize the training, validation, and test data by SentencePiece \citep{sentencepiece}, with a vocabulary size of 10,000 for the IWSLT 14 De-En dataset, and a vocabulary size of 32,000 for the WMT 14 En-Fr dataset. After tokenizing the data, we filter the training datasets, keeping only sentences that have no more than 360 tokens for IWSLT 14 De-En, and keeping only sentences that have mo nore than 200 tokens for WMT 14 En-Fr. We share the embeddings for both the encoder and the decoder Transformer, and use the same embedding matrix for softmax in the decoder. We train our models using Adam, with a learning rate linearly warming up for 4,000 steps to $1.6 \times 10^{-3}$, and then decreasing to 0 using the cosine schedule. We train for 15,000 steps on IWSLT 14 De-En, and 35,000 steps on WMT 14 En-Fr. We do \textit{not} use checkpoint averaging for decoding, which could potentially improve our results. 

When we transfer the dropout pattern found on Penn Treebank to our machine translation experiments, we keep the same hyper-parameters: \texttt{size}, \texttt{stride}, \texttt{share\_t}, and \texttt{share\_c}. Unlike the language model tasks, we do not use embedding dropout or softmax dropout. We also set the dropout rate at all steps to 0.1.

\section{\label{sec:noise_visualizaion}Visualization of Good Dropout Patterns}
\begin{figure}[htb!]
\centering
\includegraphics[width=0.5\linewidth]{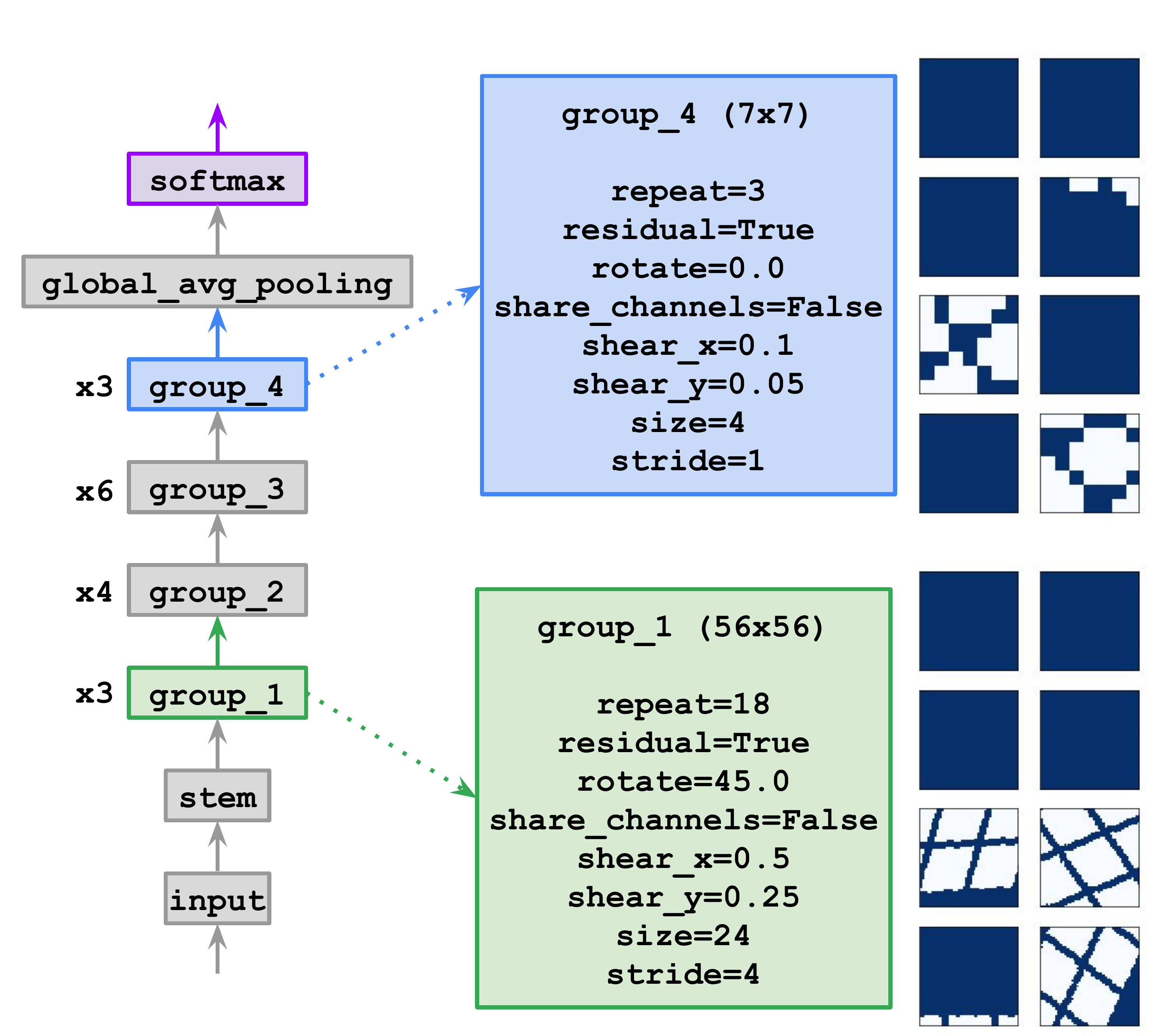}
\caption{\label{fig:resnet50_mask}The best configuration found for ResNet-50. \textbf{Left:} \methodname{} suggests that we should only apply the noise patterns at \texttt{group\_1} and \texttt{group\_4}, out of the bottleneck convolutional blocks of ResNet-50. Meanwhile, no noise is injected to \texttt{group\_2} and \texttt{group\_3}. \textbf{Middle:} detailed configurations for these two groups. \textbf{Right:} Example masks for 8 channels of the corresponding groups, where the values at white pixels are set to 0.}
\end{figure}

\begin{figure}[htb!]
\centering
\includegraphics[width=0.8\textwidth]{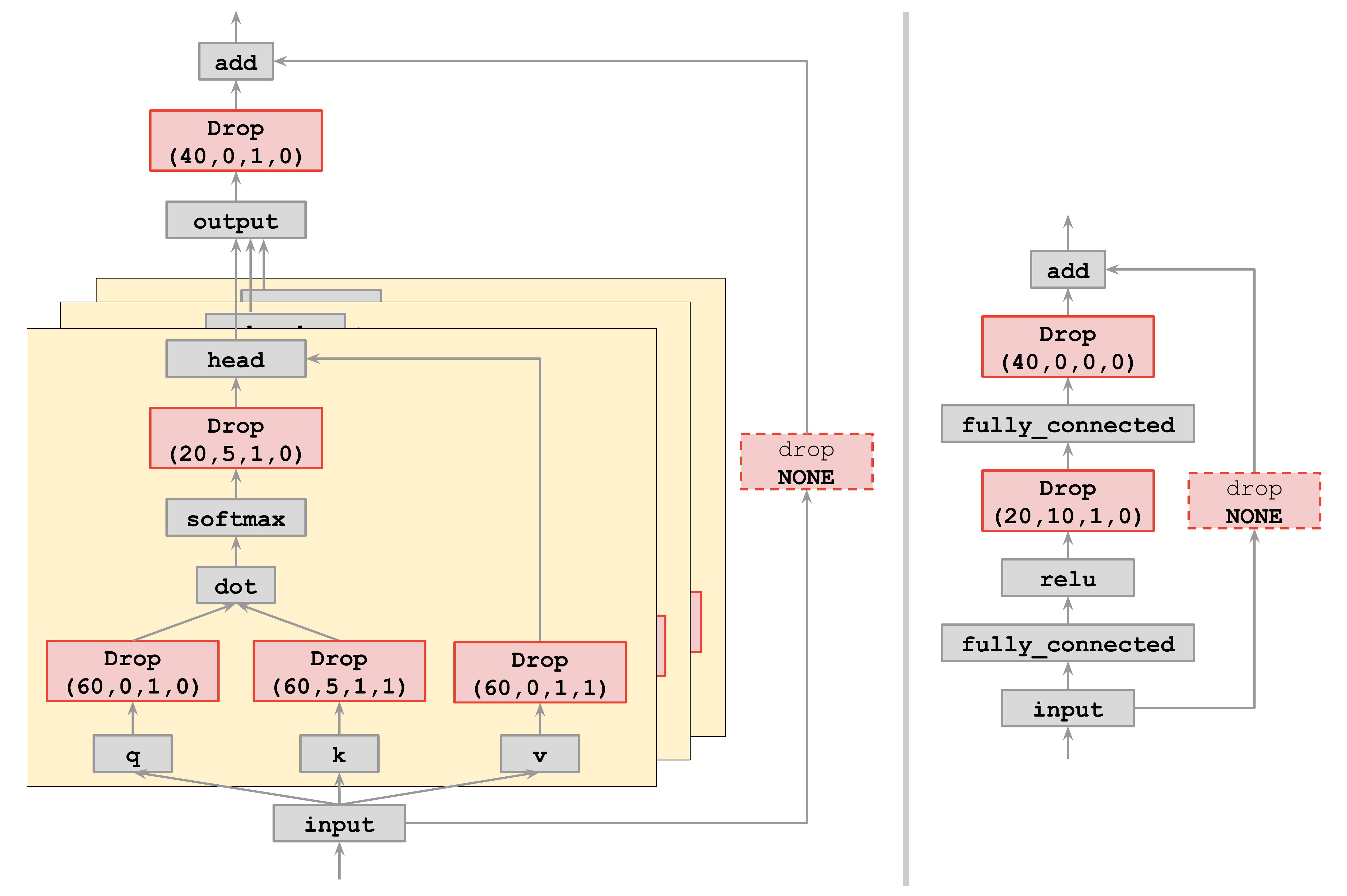}
\caption{\label{fig:transformer_mask}The best dropout pattern that \methodname{} finds for Transformer-XL on Penn Treebank. \textbf{Left:} the dropout pattern in the self-attention operation. \textbf{Right:} the dropout pattern in the positional feed-forward operation. \textbf{Meanings of the dropout pattern's hyper-parameters:} At each step where the controller can apply a dropout pattern, we specify a tuple of \texttt{(size, stride, share\_t, share\_c)}. \texttt{size} and \texttt{stride} specify how many consecutive tokens are affected by the dropout pattern, and then how many consecutive tokens are not affected by the pattern. \texttt{share\_t} means whether the dropout pattern uses the same mask at all \texttt{size} temporal steps that it affects, and \texttt{share\_c)} decides whether the pattern uses the same mask across the channel dimension. A tuple of \texttt{None} means that the controller decides to not apply any dropout pattern at the corresponding step. In this case, the controller does not apply any noise pattern on the residual branches.}
\end{figure}

\end{document}